\documentclass[lettersize,journal]{IEEEtran}
\usepackage{amsmath,amsfonts}
\usepackage{algorithmic}
\usepackage{algorithm}
\usepackage{array}
\usepackage[caption=false,font=normalsize,labelfont=sf,textfont=sf]{subfig}
\usepackage{textcomp}
\usepackage{stfloats}
\usepackage{url}
\usepackage{verbatim}
\usepackage{graphicx}
\usepackage{cite}
\makeatletter
\let\NAT@parse\undefined
\makeatother
\usepackage{hyperref}  
\usepackage{booktabs}
\usepackage{threeparttable}
\usepackage[table,xcdraw]{xcolor}

\begin{document}

\title{YOLC: You Only Look Clusters\\ for Tiny Object Detection in Aerial Images}

\author{Chenguang Liu, Guangshuai Gao, Ziyue Huang, Zhenghui Hu, Qingjie Liu, ~\IEEEmembership{Member,~IEEE}, and Yunhong Wang, ~\IEEEmembership{Fellow,~IEEE}
\thanks{This work was supported in part by the National Natural Science Foundation of China under Grant 62176017, Grant 41871283, and Grant 62301623, and in part by Henan province key science and technology research projects under Grant 232102211002. (\textit{Corresponding author: Qingjie Liu}.)}
\thanks{Chenguang Liu, Ziyue Huang, Qingjie Liu, and Yunhong Wang are with the State Key Laboratory of Virtual Reality Technology and Systems, Beihang University, Haidian, Beijing 100191, China, and also with the Hangzhou Innovation Institute, Beihang University, Hangzhou 310051, China (e-mail: liuchenguang@buaa.edu.cn; ziyuehuang@buaa.edu.cn; qingjie.liu@buaa.edu.cn; yhwang@buaa.edu.cn).

Guangshuai Gao is with the School of Electronics and Information, Zhongyuan University of Technology, Zhengzhou 450007, China (e-mail: 6911@zut.edu.cn)

Zhenghui Hu is with the Hangzhou Innovation Institute, Beihang University, Hangzhou 310051, China (e-mail: zhenghuihu2021@buaa.edu.cn).}}

\markboth{IEEE Transactions on Intelligent Transportation Systems}%
{Shell \MakeLowercase{\textit{et al.}}: A Sample Article Using IEEEtran.cls for IEEE Journals}


\maketitle

\begin{abstract}
Detecting objects from aerial images poses significant challenges due to the following factors: 1) Aerial images typically have very large sizes, generally with millions or even hundreds of millions of pixels, while computational resources are limited. 2) Small object size leads to insufficient information for effective detection. 3) Non-uniform object distribution leads to computational resource wastage.
To address these issues, we propose YOLC (You Only Look Clusters), an efficient and effective framework that builds on an anchor-free object detector, CenterNet. To overcome the challenges posed by large-scale images and non-uniform object distribution, we introduce a Local Scale Module (LSM) that adaptively searches cluster regions for zooming in for accurate detection. Additionally, we modify the regression loss using Gaussian Wasserstein distance (GWD) to obtain high-quality bounding boxes. Deformable convolution and refinement methods are employed in the detection head to enhance the detection of small objects.
We perform extensive experiments on two aerial image datasets, including Visdrone2019 and UAVDT, to demonstrate the effectiveness and superiority of our proposed approach.
\end{abstract}


\section{Introduction}
\IEEEPARstart{O}{bject} detection has witnessed remarkable progress in recent years, particularly with the rapid advancement of deep learning. Object detectors (e.g., Faster R-CNN~\cite{ren2016faster}, YOLO~\cite{redmon2016you}, and SSD~\cite{liu2016ssd}) have attained remarkable results on natural image datasets (e.g., MS COCO~\cite{lin2014microsoft}, Pascal VOC~\cite{everingham2010pascal}). However, their performance on aerial images falls short of satisfactory levels in terms of accuracy and efficiency.

\begin{figure}[t]
	\centering
	\includegraphics[width=1.0 \linewidth]{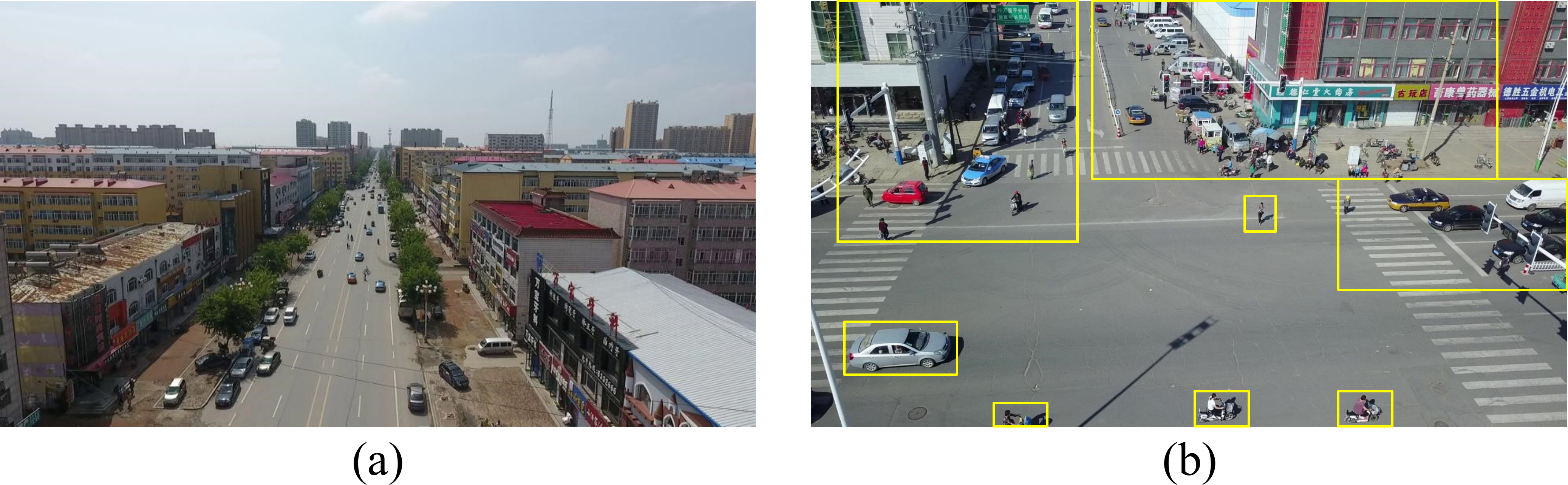}
	\caption{The challenges for the object detection in aerial images. (a) There are many small objects that are difficult to recognize for the detectors. (b) The data distribution is non-uniform.}
	\vspace{-5pt}
	\label{fig:motivation}
\end{figure}

Aerial images are typically captured by Unmanned Aerial Vehicles (UAVs), airplanes, and satellites, resulting in a bird's eye view and wide visual field that distinguishes them from natural images. Object detection in aerial images is more challenging, due to three primary reasons:
1) Aerial images are usually extremely large, surpassing the processing capabilities of current devices. As such, these images need to be resized to a smaller size or split into small crops for detection. 2) Tiny objects constitute a significant portion of aerial images, making it difficult for detectors to recognize small objects with limited resolution and visual features. In general, small objects refer to objects with an area less than 32$\times$32~\cite{lin2014microsoft}. 3) Objects in aerial images are not uniformly distributed. For instance, in Fig.~\ref{fig:motivation} (b), some cars cluster at intersections while some cars or people appear sporadically. Objects in dense regions deserve more attention, while the background should be ignored.

Early works~\cite{singh2018snip, singh2018sniper} propose novel multi-scale training strategies, such as scale normalization for image pyramids (SNIP)~\cite{singh2018snip}, and its improved version~\cite{singh2018sniper}. These methods can effectively improve the detection performance of small objects, while multi-scale training imposes substantial demands on both computational resources and memory capacity. 
Another way is to improve the image resolution or feature resolution. For instance, generative adversarial networks (GAN) can be used to compensate for the information loss of small objects~\cite{li2017perceptual}. Therefore, the gap between the feature representation of small and large objects can be narrowed; but the computational cost would be expensive.
Recently, some label-assignment based methods~\cite{xu2022nwd,xu2022rfla} aim to improve sample assignment strategies for rare small object samples. They have enhanced the detection performance of small objects, but there is still potential for improvement in terms of accuracy or efficiency.

The non-uniform distribution of small objects in high-resolution aerial images poses significant challenges to detectors, leading to reduced efficiency or accuracy on large-scale aerial images. To tackle these issues, a straightforward approach is to split the images into several crops and scale them up, as demonstrated in uniform cropping~\cite{ozge2019power}. However, this method fails to account for the non-uniform distribution of objects, and detecting all the crops still requires a substantial amount of time.


\IEEEpubidadjcol

To address the aforementioned challenges, mainstream solutions have been proposed that involve designing dedicated schemes to locate cluster regions~\cite{yang2019clustered,li2020density,ozge2019power,zhang2019fully,deng2020global}, which can subsequently be used for detection. ClusDet~\cite{yang2019clustered} employs a clustered detection network to detect clusters of objects. DMNet~\cite{li2020density} models the object distribution and generates cluster regions via density maps. These strategies exhibit promising results as the clustered regions are preserved, and the backgrounds are suppressed as much as possible. However, the independent detection of each crop reduces the inference speed. Additionally, while the above methods generate cluster regions, object distribution in some clusters is sparse, leading to little contribution to the final performance. Consequently, achieving an optimal trade-off between accuracy and efficiency is a critical problem for object detection in aerial images.

Inspired by our observations, we propose an innovative anchor-free cluster network called \textbf{Y}ou \textbf{O}nly \textbf{L}ook \textbf{C}lusters (\textbf{YOLC}) for aerial image object detection. Our model uses CenterNet~\cite{zhou2019objects} with simplicity, scalability, and high inference speed as the baseline, which directly predicts the centers of objects without relying on dedicated anchor boxes. This makes it a more appropriate option for detecting dense and small objects in aerial images. Notably, other researchers have leveraged the CenterNet framework to model remote sensing objects using centerness-aware networks~\cite{shi2022canet} or keypoints~\cite{cui2021sknet}. CenterNet~\cite{zhou2019objects} uses density maps, a.k.a heatmaps, to estimate the object locations. The heatmap provides insight into the object distribution, allowing us to distinguish between cluster regions and sparse regions. Building on this insight, we develop a Local Scale Module (LSM) that can adaptively search for cluster regions and resize them to the appropriate scale for the detectors to operate effectively.

The proposed LSM has several advantages over existing image cropping strategies~\cite{yang2019clustered,li2020density,deng2020global}. Firstly, it is an unsupervised module, which means that it can be seamlessly integrated into any keypoint-based detector without requiring an extra cluster proposal network. This simplifies the training process and reduces the overall complexity of the detection system. Secondly, the LSM is easy to implement and has low memory consumption, which makes it suitable for real-world applications. Lastly, compared with existing works~\cite{yang2019clustered,li2020density,deng2020global}, LSM generates fewer image crops and achieves excellent performance with high inference speed. This makes it a practical solution for detecting objects in large-scale aerial images.

To achieve more precise object detection, we make several improvements to CenterNet. Firstly, we modify the regression loss using Gaussian Wasserstein distance (GWD)~\cite{yang2021rethinking}, which is particularly suitable for detecting small objects but can lead to reduced performance for larger objects. To address this, we propose the GWD+$L_1$ loss, which combines the advantages of both loss functions.

Additionally, we improve the detection head by using deformable convolutions to refine the bounding box regression. We also devise a decoupled branch for the heatmap, which allows for more accurate localization of different object categories.

Overall, these improvements allow our YOLC model to achieve more precise and accurate object detection in aerial images.

In summary, the main contributions are as follows:

1) We propose a novel and efficient anchor-free object detection framework, \textbf{YOLC}, which achieves state-of-the-art performance on two aerial image datasets. Compared with many existing methods, it is simple and elegant, containing only one network, resulting in fewer parameters and higher efficiency.

2) A lightweight unsupervised local scale module (\textbf{LSM}) is introduced to adaptively search cluster regions.

3) A refined regression loss function based on GWD is designed, which is more efficient for small object detection.

4) An improved detection head is also proposed, which leverages deformable convolution for accurate bounding box regression, and a disentangled heatmap branch is devised to localize different categories of objects precisely.

5) Extensive experiments on two aerial image datasets demonstrate the proposed approach's effectiveness and superiority over state-of-the-art methods.

\section{Related Work}
General object detection methods can be broadly classified into two categories: anchor-based and anchor-free methods. Anchor-based models can be further divided into two-stage and single-stage methods, based on whether they utilize candidate proposals. Two-stage methods mainly include Faster RCNN~\cite{ren2016faster}, Mask RCNN~\cite{he2017mask}, and R-FCN~\cite{dai2016r}. On the other hand, single-stage detectors directly regress the bounding boxes and classes without a proposal stage. Representative examples of single-stage methods are YOLOs~\cite{redmon2017yolo9000,redmon2018yolov3}, SSD~\cite{liu2016ssd}, and RetinaNet~\cite{lin2017focal}. Anchor-free models, on the other hand, do not require complicated hand-crafted anchors. Some anchor-free models, such as Foverbox~\cite{kong2020foveabox} and FCOS~\cite{tian2019fcos}, predict the distances between locations to the four sides of the bounding boxes. Alternatively, other anchor-free methods only predict the corner or center points of the objects, including CornerNet~\cite{law2018cornernet}, Cornernet-lite~\cite{law2019cornernet}, and CenterNet~\cite{duan2019centernet,zhou2019objects}. While these detectors perform well in natural scenes, they often fail to obtain satisfactory results when applied to aerial images.

Small object detection is a challenging yet significant topic. Unlike general object detection, small object detection encounters issues such as information loss, noisy feature representation, low tolerance for bounding box perturbation and inadequate samples~\cite{cheng2023towards}. Cheng et al.~\cite{cheng2023towards} provide an extensive review of small object detection methods, categorizing them into various groups such as sample-oriented~\cite{xu2022rfla, kisantal2019augmentation}, scale-aware~\cite{singh2018snip,singh2018sniper,yang2022querydet,liu2018path}, super-resolution-based~\cite{li2017perceptual,bai2018sod}, context modeling~\cite{pang2019r2}, and focus-and-detect~\cite{yang2019clustered,li2020density, duan2021coarse}. However, aerial images typically contain numerous small objects and have large size. Detecting on massive backgrounds consumes computation with no effort.
Among the various methods, focus-and-detect techniques are found to be more accurate and memory-efficient, particularly outperforming super-resolution methods. 
Hence, focus-and-detect methods are widely embraced in aerial image object detection and are also referred to as image cropping-based methods.

Aerial image object detection poses unique challenges due to the wider view fields and abundance of small objects. To overcome these challenges, image cropping strategies have been widely adopted. For instance, {\"U}nel et al.~\cite{ozge2019power} demonstrated the effectiveness of uniform cropping for small object detection. R$^2$CNN~\cite{pang2019r2} and SAHI~\cite{akyon2022sahi} divide high-resolution images into small overlapping crops before performing detection on them. However, uniform cropping suffers from the drawback of many crops containing only backgrounds, leading to inefficient detection. To address this limitation, ClusDet~\cite{yang2019clustered} proposed a cluster proposal network (CPNet) that obtains clustered regions, and a scale estimation network (ScaleNet) to rescale the regions to fit the detector. DMNet~\cite{li2020density} and CDMNet~\cite{duan2021coarse} utilized density maps to detect objects and learn scale information. GLSAN~\cite{deng2020global} used a global-local fusion strategy and a progressive scale-varying network for accurate detection. In contrast to these methods, we introduce a local scale module that can adaptively search clustered regions and resize them to fit the detector, which is simple and efficient.


\section{Proposed Method}
\subsection{Preliminary}

CenterNet~\cite{zhou2019objects} is a powerful and efficient anchor-free framework for object detection. Unlike traditional methods that use anchors to predict bounding boxes, CenterNet regresses the size, orientation, pose, and keypoints of objects from their center points. This is achieved through a fully convolutional network that generates a heatmap (density map) of object centers, and then locates the centers by finding local maxima in the heatmap. Using features at these peak locations, the size of the objects can be inferred. Due to its simplicity, CenterNet~\cite{zhou2019objects} achieves extraordinary performance without relying on complex feature engineering. It is a fast and effective approach for object detection that has been widely adopted in the research community. CenterNet, as a representative anchor-free detector, employs high-resolution feature maps for predictions, making it particularly friendly and efficient towards small objects.

Density maps are powerful tools that provide information about the distribution of objects in an image. In CenterNet, density maps are utilized to locate objects. To improve the performance of the detector in detecting small objects, we upsample the feature maps to match the size of the input image using transpose convolution layers. Additionally, we propose a local scale module that leverages the heatmap to adaptively search clustered regions and resize them to fit the detector, which can further improve detection accuracy.

\subsection{You Only Look Clusters (YOLC)}

The proposed YOLC follows a similar pipeline to CenterNet~\cite{zhou2019objects}, but it is distinguished from CenterNet by utilizing a different backbone, detection head, regression manner, and loss functions. In particular, HRNet~\cite{sun2019high} is utilized as the backbone to generate high-resolution heatmaps that are more adept at detecting small objects. Additionally, due to the imbalanced distribution of objects in aerial images, a local scale module (LSM) has been designed to adaptively search clustered regions. Upon detecting the original image and crops, the refined results are directly substituted for the original image's results in dense regions. 

In the subsequent sections, we will delve into the details of the modules present in YOLC.

\begin{figure*}[!htbp]
	\centering
	\includegraphics[width=1.0 \linewidth]{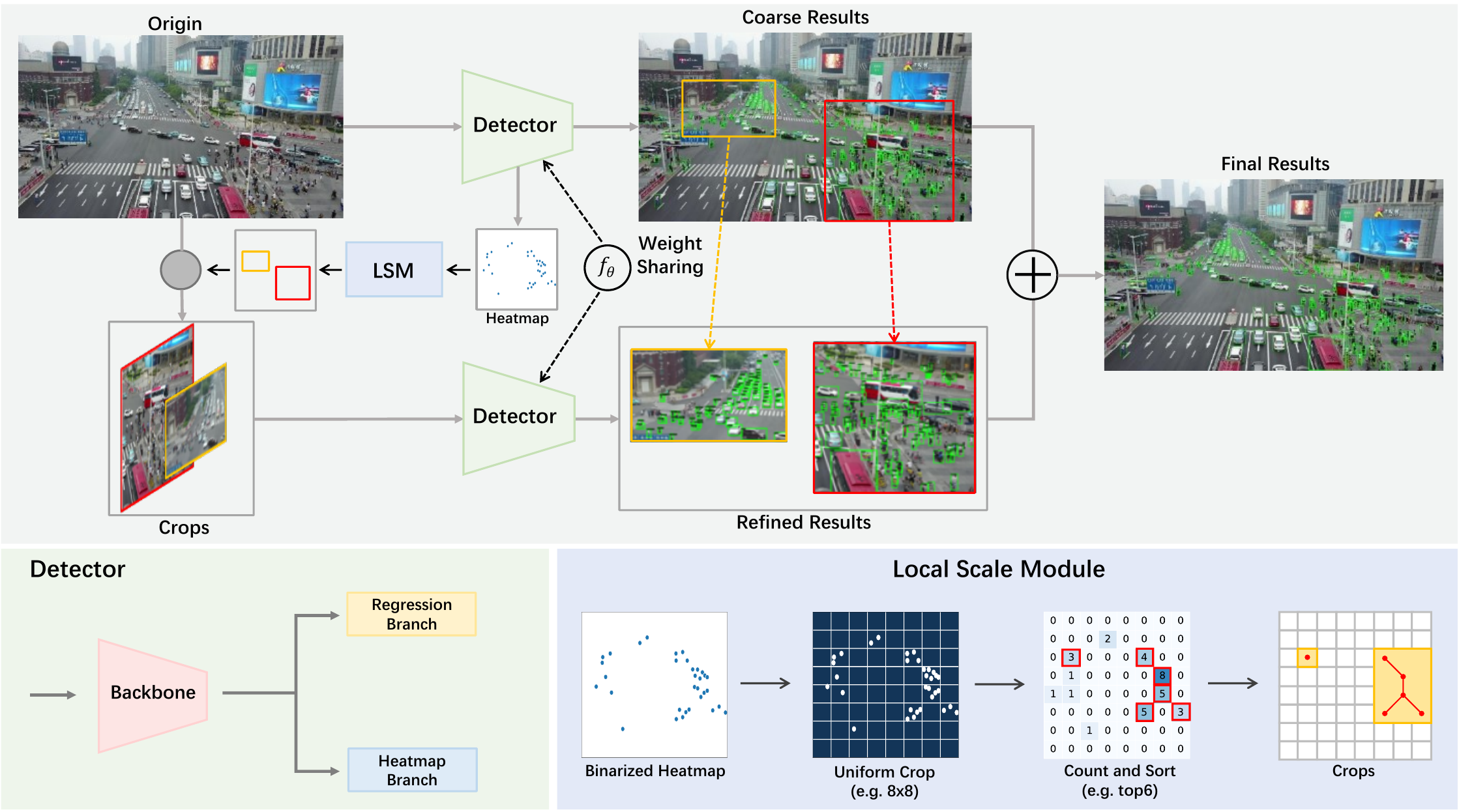}
	\caption{The YOLC framework with the local scale module. The final detection results are obtained from the combination of the image crops and global images. YOLC replaces the coarse results with refined results in the corresponding region without NMS. Noting that the global image and crops share the same detector. }
	\label{fig:LSM}
\end{figure*}
\subsection{High-resolution Heatmap}

To improve the accuracy of object detection in dense regions full of small objects, YOLC uses a higher resolution heatmap. In CenterNet, each object is modeled as a single point at the center of its bounding box, represented by a Gaussian blob in the heatmap. However, the heatmap is downsampled by a factor of $4\times$ relative to the input image. This downsampling can result in small objects collapsing into only a few or even a single point in the heatmap, making it difficult to accurately locate their centers.

To address this issue, YOLC employs a modified pipeline that uses a higher resolution heatmap. Specifically, we add one convolution layer and two transposed convolution layers to scale up the heatmap to the same size as the input image. This allows us to capture more detailed information about small objects, which in turn leads to more accurate object detection in dense regions.

Applying a Gaussian filter before decoding helps to reduce the false-positive predictions in CenterNet. The filter smooths the heatmap and suppresses the multiple peaks around the objects. This approach helps to improve the localization accuracy of the objects and reduces the chances of misclassification.

\subsection{Local Scale Module}
\label{subsec:lsm}

The region of interest proposal methods are a crucial component of cropping-based object detection models. However, in aerial images, objects such as vehicles and pedestrians tend to gather in a few clustered regions. Most regions in the images are background and do not require detection. Moreover, the limited resolution of dense regions can lead to a significant drop in detection performance. Existing cropping-based methods, such as DMNet~\cite{li2020density}, produce many crops or use additional networks like ClusDet~\cite{yang2019clustered}, resulting in low detection speeds and increased model parameters.

To address these issues, we propose a local scale module (LSM) that can adaptively locate the clustered regions. Our LSM is inspired by AutoScale~\cite{xu2021autoscale}, but we make several modifications to make it suitable for aerial images. First, instead of only searching for a single maximum clustered region, our LSM can locate the top-K dense regions by sorting the densities in each grid. This is important since aerial images often have multiple clustered regions. Second, AutoScale is designed for crowd counting and localization~\cite{gao2020feature, gao2023dacc}, which only works in scenes with single-class objects. However, in aerial images, there are multiple object categories.

We also note that UCGNet~\cite{liao2021unsupervised} uses clustering methods such as DBSCAN and K-Means to generate image crops from dense regions. However, the crops generated by UCGNet still have large size and do not consider the differences in density between different crops.

In the following, we will provide more details on our proposed LSM, and its structure is illustrated in Fig.~\ref{fig:LSM}.

To generate candidate regions, the first step is to generate a binary image indicating the existence of objects. This is done by obtaining the heatmap from the heatmap branch for each image, and then using an empirical threshold to binarize the heatmap and generate the location mask. The binarized heatmap is then divided into several grids (e.g., 16$\times$10), and the top-K (e.g., K=15) dense grids are selected as the candidate regions. For VisDrone dataset, the best hyperparameters are Grid= (16, 10) and top-K=50. To ensure complete objects are captured, eight-neighbor connected dense grids are combined into larger candidate regions, similar to~\cite{li2020density}. These regions are then enlarged by 1.2 times to avoid truncation. 

Finally, $k$ image patches are obtained by cropping dense regions from the original image and resizing them to fit the detector. The complete algorithm is illustrated in Algorithm~\ref{alg:LSM}.

To accelerate detection and achieve higher performance improvement, we aim to generate fewer crops. We set $k$=2 for each image and focus on larger dense regions, as conducting fine detection on larger crops can bring higher performance promotion. In subsequent experiments, it was observed that the performance tends to saturate with k increasing from 3. Therefore, LSM takes advantage of very few high-quality crops to detect precisely, showing a good balance between detection speed and accuracy. Furthermore, LSM is an unsupervised module that can be easily integrated into any keypoint-based detector.

\begin{figure}[!t]
        \begin{algorithm}[H]
            \caption{Local Scale Module Algorithm}
            \label{alg:LSM}
            \begin{algorithmic}[1]
                \REQUIRE Binarized heatmap $M$ with height $H$ and width $W$, the number of crops $k$
                \ENSURE Bounding boxes of cluster crops $C=\{c_i\}_{i=1}^{k}$  
                \STATE {$C\gets\{\}$}
                \STATE {$w=\frac{W}{16}, h=\frac{H}{10}$}
                \STATE {$B[1...16, 1...10]\gets\{0,...,0\}$}
                \FOR {$i=1$ to 16}
                        \FOR {$j=1$ to 10}
                        \STATE {$B[i, j] = {\rm sum} (M[w*(i-1):w*i, h*(j-1):h*j])$}
                        \ENDFOR
                \ENDFOR
                \STATE {$Grid\gets{\rm top-K}(B, topK)$}
                \STATE {$C\gets{\rm eight\mbox{-}connected}(Grid)$}
                \STATE {$C\gets{\rm sort\mbox{-}by\mbox{-}area}(C)$}
                \IF {${\rm len}(C)>k$}
                        \STATE {$C\gets C[1...k]$}
                \ENDIF
                \STATE {$C\gets {\rm enlarged}(C)$}
                \STATE {\textbf{return} $C$}
            \end{algorithmic}
        \end{algorithm}
    \end{figure}

\subsection{Loss Function}
\label{subsec:loss}
CenterNet~\cite{zhou2019objects} integrates three losses to optimize the overall network,
\begin{equation}
L_{det}=L_{k}+\lambda_{size}L_{size}+\lambda_{off}L_{off}
\end{equation}
where $L_{k}$ is the modified focal loss in CenterNet~\cite{zhou2019objects}. $L_{off}$ is the center point offset loss. $L_{size}$ is the size regression loss. $\lambda_{size}$ and $\lambda_{off}$ are set as 0.1 and 1 by default.

CenterNet's current approach of using $L_1$ loss, a.k.a mean absolute error (MAE), for size regression has limitations. Specifically, it is sensitive to object size variations and individually regresses the height and width of an object, leading to the loss ignoring the correlation between the coordinates of each object. This stands in contrast to IoU-based losses that optimize these coordinates as a whole, providing more robust predictions. Moreover, the use of $L_1$ loss can lead to inconsistencies between the metric and regression loss, as a smaller training loss does not guarantee higher performance~\cite{yang2021rethinking}. Finally, some predicted bounding boxes with the same $L_1$ loss may correspond to different IoUs between each one and the matched ground truth, further highlighting the limitations of $L_1$ loss for size regression.

To address this issue, we introduce a metric based on Gaussian Wasserstein distance (GWD)~\cite{yang2021rethinking} to measure the similarity between two bounding boxes. Specifically, given a bounding box $\mathcal{B}(x, y, h, w)$, where $(x,y)$ represent the center coordinates and $w$ and $h$ denote width and height, respectively. We first convert it into a 2-D Gaussian distribution.
Following GWD, the converted Gaussian distribution is formulated as follows:

\begin{equation}
f(\mathbf{x} \mid \boldsymbol{\mu}, \boldsymbol{\Sigma})=\frac{\exp \left(-\frac{1}{2}(\mathbf{x}-\boldsymbol{\mu})^\mathsf{T} \boldsymbol{\Sigma}^{-1}(\mathbf{x}-\boldsymbol{\mu})\right)}{2 \pi|\boldsymbol{\Sigma}|^{\frac{1}{2}}}
\end{equation}
where $\boldsymbol{\mu}$ and $\boldsymbol{\Sigma}$ stand for the mean vector and the co-variance matrix of Gaussian distribution, respectively. The $\boldsymbol{\mu}$ and $\boldsymbol{\Sigma}$ are: 

\begin{equation}
\begin{array}{l}
\boldsymbol{\mu}=(x, y)^{T},
\boldsymbol{\Sigma}=\left(\begin{array}{cc}
\omega^{2} / 4 & 0 \\
0 & h^{2} / 4
\end{array}\right) \\
\end{array}
\end{equation}

According to Optimal Transport Theory, the Wasserstein distance between two distribution $\mu$ and $\nu$ can be computed as:
\begin{equation}
\mathbf{W}(\mu ; \nu):=\inf \mathbb{E}\left(\|\mathbf{X}-\mathbf{Y}\|_{2}^{2}\right)^{1 / 2}
\end{equation}
where the inferior runs across over all random vectors $\left (\mathbf{X},\mathbf{Y}\right)$ of $\mathbb{R}^{n} \times \mathbb{R}^{n}$ with $\mathbf{X} \sim \mu$ and $\mathbf{Y} \sim \nu$. Therefore, given two 2D Gaussian distribution $\mathcal{N}(\boldsymbol{\mu}_{1}, \boldsymbol{\Sigma}_{1})$ and $\mathcal{N}(\boldsymbol{\mu}_{2}, \boldsymbol{\Sigma}_{2})$, the Wasserstein distance is calculated as:
\begin{equation}
\label{eq:GWD}
\mathbf{W}^{2} = \left\|\boldsymbol{\mu}_{1}-\boldsymbol{\mu}_{2}\right\|_{2}^{2}+\mathbf{Tr}\left(\boldsymbol{\Sigma}_{1}+\boldsymbol{\Sigma}_{2}-2\left(\boldsymbol{\Sigma}_{1}^{1 / 2} \boldsymbol{\Sigma}_{2} \boldsymbol{\Sigma}_{1}^{1 / 2}\right)^{1 / 2}\right)
\end{equation}

It is worth noting that we have:
\begin{equation}
\mathbf{Tr}\left( \left(\boldsymbol{\Sigma}_{1}^{1 / 2} \boldsymbol{\Sigma}_{2} \boldsymbol{\Sigma}_{1}^{1 / 2}\right)^{1 / 2}\right)=\mathbf{Tr}\left( \left(\boldsymbol{\Sigma}_{2}^{1 / 2} \boldsymbol{\Sigma}_{1} \boldsymbol{\Sigma}_{2}^{1 / 2}\right)^{1 / 2}\right)
\end{equation}

Since the detection task use horizontal bounding boxes as the ground truth, we have $\Sigma_{1} \Sigma_{2}=\Sigma_{2} \Sigma_{1}$. Then Eq.(\ref{eq:GWD}) can be rewritten as follows~\cite{yang2021rethinking}:

\begin{equation}
\label{eq:wassertein}
\begin{aligned}
\mathbf{W}^{2} &=\left\|\boldsymbol{\mu}_{1}-\boldsymbol{\mu}_{2}\right\|_{2}^{2}+\left\|\boldsymbol{\Sigma}_{1}^{1 / 2}-\boldsymbol{\Sigma}_{2}^{1 / 2}\right\|_{F}^{2} \\
&=\left(x_{1}-x_{2}\right)^{2}+\left(y_{1}-y_{2}\right)^{2}+\frac{\left(w_{1}-w_{2}\right)^{2}+\left(h_{1}-h_{2}\right)^{2}}{4} \\
&=l_{2} \text {-norm }\left(\left[x_{1}, y_{1}, \frac{w_{1}}{2}, \frac{h_{1}}{2}\right]^\mathsf{T},\left[x_{2}, y_{2}, \frac{w_{2}}{2}, \frac{h_{2}}{2}\right]^\mathsf{T}\right)
\end{aligned}
\end{equation}
where $\|\cdot\|_{F}$ indicates the Frobenius norm.

Only using the above loss (Eq. (\ref{eq:wassertein})) may be sensitive to large errors, therefore, a non-linear function is performed and the loss is converted into an affinity measure $\frac{1}{\tau+f\left(\mathbf{W}^{2}\right)}$. Following ~\cite{rezatofighi2019generalized,zheng2020distance}, the final GWD-based loss is expressed as~\cite{yang2021rethinking}:
\begin{equation}
L_{gwd}=1-\frac{1}{\tau+f\left(\textbf{W}^{2}\right)}, \quad \tau \geq 1
\end{equation}
where $f(\cdot)$ is the non-linear function to make the Wasserstein distance more smooth and easily to optimize the model. We define $f(x)=\ln(x+1)$. $\tau$ is a modulated hyperparmeter, which is empirically set as 1.

Since the GWD-based loss has taken the center offset into consideration, we modify the overall detection loss as follows:
\begin{equation}
L_{det}=L_{k}+\lambda_{gwd} L_{gwd}
\end{equation}
where $\lambda_{gwd}$ is a trade-off hyperparameter, which is set as 2.

According to the following experiment, replacing $L_1$ loss with GWD loss led to a significant improvement in detecting small objects. However, detection of medium and large objects did not perform as well as before. To investigate the sensitivity of GWD loss to object size, we computed the gradient of GWD loss:

\begin{equation}
\label{eq:gradient}
\nabla_\mathbf{W} L_{gwd}=\frac{2\mathbf{W}}{(1+\mathbf{W}^{2})(1+\log(1+\mathbf{W}^{2}))^{2}}
\end{equation}

Regarding large objects, we observed that the Wasserstein distance $\mathbf{W}$ tends to be high during the early phase of model training, which can lead to a gradient close to 0 and therefore affect the detection of medium and large objects. To address this issue, we propose to combine the GWD loss with $L_1$ loss. Specifically, we modify the loss function as follows:

\begin{equation}
L_{det}=L_{k}+\lambda_{gwd} L_{gwd}+\lambda_{l1} L_{1}
\end{equation}

where we set $\lambda_{gwd}$ to 2, $\lambda_{l1}$ to 0.5.

By introducing $L_1$ loss, the model can still learn the regression of the size of large objects, which helps to overcome the above issue. This modified loss function can significantly improve the performance of YOLC on detecting objects of various sizes.

\subsection{Improved Detection Head}

\begin{figure}[t]
	\centering
	\includegraphics[width=1.0 \linewidth]{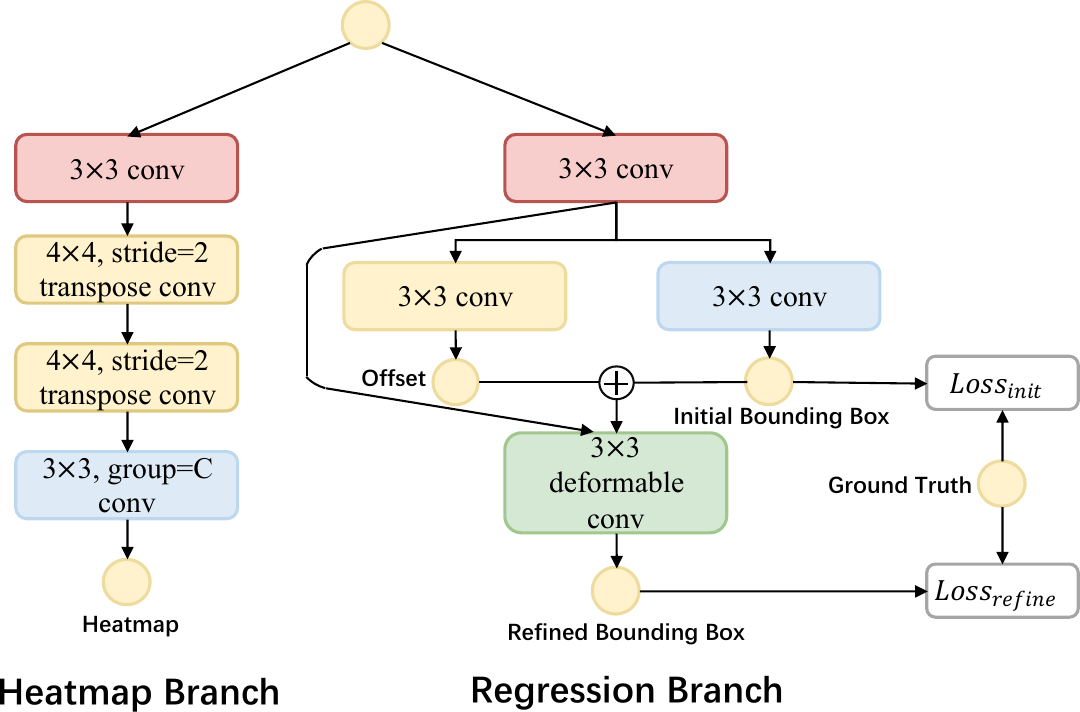}
	\caption{The structure of YOLC detection head with the regression branch and heatmap branch.}
	\vspace{-5pt}
	\label{fig:head}
\end{figure}

To improve the detection of small objects in aerial images, the regression branch is enhanced with deformable convolution, as it can adaptively adjust the sampling locations in the convolution operation to better capture small details. Additionally, to better capture the fine details of different categories of objects, the heatmap branch is decoupled into multiple sub-branches, each responsible for predicting the heatmaps of a specific object category. This not only reduces the computational burden of predicting all heatmaps simultaneously but also allows the network to focus on learning the distinct characteristics of each category, improving the overall performance of the detector. The structure of the detection head with these improvements is illustrated in Figure~\ref{fig:head}.

\textbf{Regression branch}. In traditional object detectors, point features are used to recognize objects, resulting in inaccurate predictions. To overcome this limitation, we adopt deformable convolution in the regression branch. Unlike regular convolution that operates on a fixed 3$\times$3 grid, deformable convolution dynamically adjusts the sampling points to fit the object's pose and shape. Specifically, a regular 3$\times$3 grid has 9 sampling points, defined as \{(-1, -1), (-1, 0), …, (1, 1)\}, and deformable convolutions estimate offsets from the regular sampling points to the dynamic sampling points. 

In our approach, we predict coarse bounding boxes and offset fields in parallel. To ensure that the convolution covers the whole object, we constrain the offsets within the coarse bounding boxes and apply a deformable convolution layer on them. This produces refined bounding boxes with more accurate locations and sizes. When we calculate loss, the loss between the initial bounding boxes and the ground truth is denoted by $Loss_{initial}$. In the same way, we can calculate $Loss_{refine}$ about the refined bounding boxes (see Fig.~\ref{fig:head}).

\textbf{Heatmap branch}. We improve the heatmap branch by introducing a disentangled structure, inspired by the approach used in YOLOX~\cite{ge2021yolox} and DEKR~\cite{geng2021bottom}. The model now predicts heatmaps for different object categories in parallel, which allows for more efficient and accurate detection. Specifically, the heatmap branch includes a standard convolution, two transpose convolutions, and a group convolution with $C$ groups, where each group corresponds to a specific object category.

\section{Experiments and analysis}
This section presents the datasets, evaluation metrics, and implementation details used in our experiments. We then perform ablation studies to verify the effectiveness of the proposed modules, backbones, and loss functions. Finally, we conduct comparison experiments from both qualitative and quantitative perspectives to demonstrate the superiority of our approach compared with state-of-the-art methods.

\subsection{Datasets}

We evaluate our approach on two publicly available aerial image datasets: VisDrone 2019~\cite{zhu2018visdrone} and UAVDT~\cite{du2018unmanned}. The details of these two datasets are as follows:

\textbf{VisDrone}~\cite{zhu2018visdrone} comprises 10,209 high-resolution images (about 2000$\times$1500 pixels) with ten categories: pedestrian, people, bicycle, car, van, truck, tricycle, awning-tricycle, bus, and motor. The dataset contains 6,471 images for training, 548 for validation, and 3,190 for testing. Since the evaluation server has been closed, following \cite{yang2019clustered,li2020density}, we use the validation set for evaluation.

\textbf{UAVDT}~\cite{du2018unmanned} consists of 38,327 images with an average resolution of 1,080$\times$540 pixels. The dataset includes three categories, namely car, bus, and truck. The dataset contains 23,258 images for training and 15,069 images for testing.

\subsection{Evaluation Metrics}
Following the evaluation protocol on MS COCO~\cite{lin2014microsoft} dataset, we use $AP$, $AP_{50}$, and $AP_{75}$ as our evaluation metrics. $AP$ represents the average precision over all categories, while $AP_{50}$ and $AP_{75}$ denote the average precision calculated at the IoU thresholds of 0.5 and 0.75 over all categories. In addition, we report the average precision for each object category to measure the performance of each class. Furthermore, to measure the performance of different object scales, we adopt three metrics, namely $AP_{small}$, $AP_{medium}$, and $AP_{large}$. Finally, we also report the efficiency of our approach by measuring the processing time of one original image per GPU.

\subsection{Implementation Details}

Our proposed approach is implemented using the open-source MMDetection toolbox\footnote{https://github.com/open-mmlab/mmdetection}. CenterNet~\cite{zhou2019objects} with Hourglass-104~\cite{newell2016stacked} backbone is employed as the baseline method. The model is trained for 160 epochs using the SGD optimizer with a momentum of 0.9 and weight decay of 0.0001. It is run on the GeForce RTX 2080 Ti GPU platform with a batch size of 2. The initial learning rate is set to 0.01 with a linear warm-up. The input resolution is set to $1024\times 640$ for both datasets.

\begin{table*}[!ht]
  \vspace{-0.3cm}
  \caption{Performance comparison on VisDrone~\cite{zhu2018visdrone}. ``o", ``c", ``ca", and ``aug" respectively stand for the original validation set, evenly image partition (EIP) cropped images, cluster-aware cropped images, and augmented images. ``$*$" denotes the multi-scale inference. ``$\dag$" denotes optimization through hyperparameter tuning.}
	\begin{center}
		\resizebox{\textwidth}{!}{
		\begin{tabular}{c|ccccc|ccc|ccc|c}
        \toprule[1pt]
        Method &Year$\&$Venue &Baseline &Backbone &test data &\#img &$AP$ &$AP_{50}$ &$AP_{75}$ &$AP_{small}$ &$AP_{medium}$ &$AP_{large}$ &s/img (GPU) \\  \midrule
        FRCNN~\cite{ren2016faster}+FPN~\cite{lin2017feature} &2015 NIPS&-- &ResNet50 &o &548 &21.4 &40.7 &19.9 &11.7 &33.9 &54.7 &0.055   \\
        FRCNN~\cite{ren2016faster}+FPN~\cite{lin2017feature} &2015 NIPS&-- &ResNet101 &o &548 &21.4 &40.7 &20.3 &11.6 &33.9 &54.9 &0.074   \\
        FRCNN~\cite{ren2016faster}+FPN~\cite{lin2017feature} &2015 NIPS&-- &ResNeXt101 &o &548 &21.8 &41.8 &20.1 &11.9 &34.8 &55.5 &0.156   \\
        FRCNN~\cite{ren2016faster}+FPN~\cite{lin2017feature}+EIP &2015 NIPS&-- &ResNet50 &c &3,288 &21.1 &44.0 &18.1 &14.4 &30.9 &30.0 &0.330   \\
        FRCNN~\cite{ren2016faster}+FPN~\cite{lin2017feature}+EIP &2015 NIPS&-- &ResNet101 &c &3,288 &23.5 &46.1 &21.1 &17.1 &33.9 &29.1 &0.444   \\
        FRCNN~\cite{ren2016faster}+FPN~\cite{lin2017feature}+EIP &2015 NIPS&-- &ResNeXt101 &c &3,288 &24.4 &47.8 &21.8 &17.8 &34.8 &34.3 &0.936   \\ \midrule 

        ClusDet~\cite{yang2019clustered} &2019 ICCV &FRCNN+FPN &ResNet50 &o+ca &2,716 &26.7 &50.6 &24.7 &17.6 &38.9 &51.4 &0.273 \\
        ClusDet~\cite{yang2019clustered} &2019 ICCV &FRCNN+FPN &ResNet101 &o+ca &2,716 &26.7 &50.4 &25.2 &17.2 &39.3 &54.9 &0.366 \\
        ClusDet~\cite{yang2019clustered} &2019 ICCV &FRCNN+FPN &ResNeXt101 &o+ca &2,716 &28.4 &53.2 &26.4 &19.1 &40.8 &54.4 &0.773\\
        ClusDet$^{*}$~\cite{yang2019clustered} &2019 ICCV &FRCNN+FPN &ResNeXt101 &o+ca &2,716 &32.4 &56.2 &31.6 &-- &-- &-- &0.773\\ \midrule 

        DREN~\cite{zhang2019fully} &2019 ICCVW&MRCNN+FPN &ResNeXt101 &-- &-- &27.1 &-- &-- &-- &-- &-- &-- \\
        DREN~\cite{zhang2019fully} &2019 ICCVW&MRCNN+FPN &ResNeXt152 &-- &-- &30.3 &-- &-- &-- &-- &-- &-- \\ \midrule 

        DMNet~\cite{li2020density} &2020 CVPRW &FRCNN+FPN &ResNet50 &o+ca &2,736 &28.2 &47.6 &28.9 &19.9 &39.6 &55.8 &0.29 \\
        DMNet~\cite{li2020density} &2020 CVPRW &FRCNN+FPN &ResNet101 &o+ca &2,736 &28.5 &48.1 &29.4 &20.0 &39.7 &57.1 &0.36 \\
        DMNet~\cite{li2020density} &2020 CVPRW &FRCNN+FPN &ResNeXt101 &o+ca &2,736 &29.4 &49.3 &30.6 &21.6 &41.0 &56.7 &0.61 \\ \midrule 

        CDMNet~\cite{duan2021coarse} &2021 ICCVW &FRCNN &ResNet50 &ca &2,170 &29.2 &49.5 &29.8 &20.8 &40.7 &41.6 &-- \\
        CDMNet~\cite{duan2021coarse} &2021 ICCVW &FRCNN &ResNet101 &ca &2,170 &29.7 &50.0 &30.9 &21.2 &41.8 &42.9 &-- \\
        CDMNet~\cite{duan2021coarse} &2021 ICCVW &FRCNN &ResNeXt101 &ca &2,170 &30.7 &51.3 &32.0 &22.2 &42.4 &44.7 &-- \\ \midrule 

        AMRNet~\cite{wei2020amrnet} &2020 Arxiv &FRCNN+FPN &ResNet50 &o+aug &-- &31.7 &-- &-- &23.0 &43.4 &58.1  &--\\
        AMRNet~\cite{wei2020amrnet} &2020 Arxiv &FRCNN+FPN &ResNet101 &o+aug &-- &31.7 &-- &-- &29.0 &45.5 &\textbf{60.9} &--\\
        AMRNet~\cite{wei2020amrnet} &2020 Arxiv &FRCNN+FPN &ResNeXt101 &o+aug &-- &32.1 &-- &-- &23.2 &43.9 &60.5 &--\\ \midrule

        GLSAN~\cite{deng2020global} &2021 TIP &FRCNN+FPN &ResNet50 &o+ca &-- &30.7 &55.4 &30.0 &-- &-- &-- &-- \\
        GLSAN~\cite{deng2020global} &2021 TIP &FRCNN+FPN &ResNet101 &o+ca &-- &30.7 &55.4 &30.0 &-- &-- &-- &-- \\
        GLSAN~\cite{deng2020global} &2021 TIP &Cascade R-CNN &ResNet50 &o+ca &-- &32.5 &55.8 &30.0 &-- &-- &-- &-- \\  \midrule

        AFSM~\cite{gong2020towards} &2020 Arxiv &CenterNet &CBResNet50 &-- &-- &33.95 &60.46 &32.69 &-- &-- &-- &-- \\
        AFSM$^{*}$~\cite{gong2020towards} &2020 Arxiv &CenterNet &CBResNet50 &-- &-- &37.62 &65.41 &37.06 &-- &-- &-- &-- \\
        AFSM~\cite{gong2020towards} &2020 Arxiv &CenterNet &CBResNet50+DCN &-- &-- &35.43 &61.88 &34.60 &-- &-- &-- &-- \\
        AFSM$^{*}$~\cite{gong2020towards} &2020 Arxiv &CenterNet &CBResNet50+DCN &-- &-- &39.48 &\textbf{66.98} &39.45 &-- &-- &-- &-- \\    \midrule 

        HRDNet~\cite{liu2021hrdnet} &2021 ICME &Cascade R-CNN &ResNet18+101 &-- &-- &28.33 &49.25 &28.16 &-- &-- &-- &--  \\
        HRDNet$^{*}$~\cite{liu2021hrdnet} &2021 ICME &Cascade R-CNN &ResNeXt50+101 &-- &-- &35.51 &62.00 &35.13 &-- &-- &-- &--  \\  \midrule

        FCOS+SAHI~\cite{akyon2022sahi} &2022 ICIP &FCOS &-- &ca &-- &-- &38.5 &-- &-- &-- &-- &--  \\
        VFNet+SAHI~\cite{akyon2022sahi} &2022 ICIP &VFNet &-- &ca &-- &-- &42.2 &-- &-- &-- &-- &--  \\
        TOOD+SAHI~\cite{akyon2022sahi} &2022 ICIP &TOOD &-- &ca &-- &-- &43.5 &-- &-- &-- &-- &--  \\  \midrule

        QueryDet~\cite{yang2022querydet} &2022 CVPR &RetinaNet &ResNet50 &o &584 &28.32 &48.14 &28.75 &-- &-- &-- &0.364  \\
        CEASC~\cite{du2023ceasc} &2023 CVPR &RetinaNet &ResNet50 &o &584 &20.8 &35.0 &27.7 &-- &-- &-- &0.070  \\  
        CEASC~\cite{du2023ceasc} &2023 CVPR &GFL &ResNet18 &o &584 &28.7 &50.7 &28.4 &-- &-- &-- &0.046  \\  \midrule

        CenterNet (Baseline)~\cite{zhou2019objects} &2019 Arxiv &-- &Hourglass104 &o &548 &27.8 &47.9 &27.6 &21.3 &42.1 &49.8 &0.215 \\
        CenterNet~\cite{zhou2019objects} &-- &CenterNet &HRNet &o &548 &29.1 &49.2 &30.1 &22.3 &43.9 &52.4 &0.179 \\
        YOLC (k=0) &-- &CenterNet &HRNet &o &584 &35.5 &59.3 &35.9 &26.9 &47.8 &59.4 &0.179 \\
        YOLC (k=1) &-- &CenterNet &HRNet &o+ca &1168 &37.7 &61.7 &39.0 &30.2 &48.4 &56.4 &0.358 \\
        YOLC (k=2) &-- &CenterNet &HRNet &o+ca &1644 &37.8 &61.7 &39.4 &30.5 &48.3 &55.4 &0.537 \\
        YOLC (k=3) &-- &CenterNet &HRNet &o+ca &2192 &37.8 &61.6 &39.4 &30.6 &48.1 &54.1 &0.716 \\
        
        YOLC (k=3)\dag &-- &CenterNet &HRNet &o+ca &2192 &38.3 &62.3 &40.1 &31.7 &48.0 &46.5 &0.956 \\
        \midrule 

        CenterNet &-- &CenterNet &ResNet50 &o &584 &23.9 &45.5 &22.1 &14.4 &36.3 &44.9 &0.073 \\
        YOLC (k=0) &-- &CenterNet &ResNet50 &o &584 &28.9 &51.4 &28.3 &20.1 &41.6 &47.3 &0.136 \\
        YOLC (k=0) &-- &CenterNet &ResNet101 &o &584 &29.7 &52.4 &29.2 &20.1 &43.1 &52.4 &0.158 \\       \midrule

        YOLC (k=2) &-- &CenterNet &ResNet50 &o+ca &1168 &31.8 &55.0 &31.7 &24.7 &42.3 &45 &0.441 \\
        YOLC (k=2) &-- &CenterNet &ResNeXt101 &o+ca &1644 &33.7 &57.4 &33.8 &26.1 &44.9 &52.8 &0.543 \\
        YOLC (k=2)$^{*}$ &-- &CenterNet &ResNeXt101 &o+ca &1644 &36.3 &60.1 &37.4 &28.9 &47.5 &51.8 &0.543 \\    \midrule 
        \textbf{YOLC (k=2)$^{*}$} &-- &CenterNet &HRNet &o+ca &1644 &\textbf{39.6} &63.7 &\textbf{41.6} &32.8 &\textbf{49.4} &56.7 &0.537 \\
        \textbf{YOLC (k=3)$^{*}$} &-- &CenterNet &HRNet &o+ca &2192 &\textbf{39.6} &63.7 &\textbf{41.6} &\textbf{32.9} &49.3 &56.2 &0.716 \\

        \bottomrule[1pt]
		\end{tabular}}
	\end{center}
\label{tab:visdrone}
\end{table*}

\subsection{Experimental Results}

Table~\ref{tab:visdrone} and Table~\ref{tab:uavdt} present the quantitative comparisons with state-of-the-art methods on VisDrone~\cite{zhu2018visdrone} and UAVDT~\cite{du2018unmanned} datasets, respectively. The results show that our proposed model consistently outperforms other methods on both datasets. Notably, on VisDrone~\cite{zhu2018visdrone}, general object detectors such as Faster R-CNN~\cite{ren2016faster} and CenterNet~\cite{zhou2019objects} underperform due to the large number of small object instances and non-uniform data distribution in aerial images. However, the anchor-free detection model (i.e., CenterNet~\cite{zhou2019objects}) performs better than anchor-based detection methods (i.e., Faster R-CNN~\cite{ren2016faster}), which is consistent with the analysis mentioned earlier.

Our proposed YOLC outperforms existing methods, such as ClusDet~\cite{yang2019clustered} and DMNet~\cite{li2020density}, even without using LSM. However, the performance is further boosted after incorporating LSM. By using optimization schemes, we achieved the best performance with 38.3 AP. We also evaluated the detection results using multi-scale testing, which led to a further 1.8\% improvement in AP. Notably, our proposed YOLC significantly improved the detection performance of small objects, while the improvements in medium and large objects were not as noticeable. This is acceptable since small objects are the majority in aerial images, and improving their detection can significantly boost overall performance.

Additionally, our proposed method achieved the best detection results with relatively fewer processed images ($\#$ img), higher inference speed (s/img), and a simpler framework with fewer parameters. This indicates the effectiveness and efficiency of our approach.

\begin{table*}[!htbp]
  \vspace{-0.3cm}
  \caption{Performance comparison on UAVDT~\cite{du2018unmanned}.}
	\begin{center}
		\begin{tabular}{c|ccc|ccc|ccc}
        \toprule[1pt]
        Method  &Baseline &Backbone &\#img &$AP$ &$AP_{50}$ &$AP_{75}$ &$AP_{small}$ &$AP_{medium}$ &$AP_{large}$ \\  \midrule
        R-FCN~\cite{dai2016r} &-- &ResNet50  &15,069 &7.0 &17.5 &3.9 &4.4 &14.7 &12.1 \\
        SSD~\cite{liu2016ssd} &-- &-- &15,069 &9.3 &21.4 &6.7 &7.1 &17.1 &12.0 \\
        FRCNN~\cite{ren2016faster} &-- &--  &15,069 &5.8 &17.4 &2.5 &3.8 &12.3 &9.4 \\
        FRCNN~\cite{ren2016faster}+FPN~\cite{lin2017feature} &-- &-- &15,069 &11.0 &23.4 &8.4 &8.1 &20.2 &26.5 \\
        FRCNN~\cite{ren2016faster}+FPN~\cite{lin2017feature}+EIP  &-- &-- &60,276 &6.6 &16.8 &3.4 &5.2 &13.0 &17.2 \\ \midrule
        ClusDet~\cite{yang2019clustered}  &FRCNN+FPN &ResNet50 &25,427 &13.7 &26.5 &12.5 &9.1 &25.1 &31.2 \\ 
        DMNet~\cite{li2020density}  &FRCNN+FPN &ResNet50 &32,764 &14.7 &24.6 &16.3 &9.3 &26.2 &35.2  \\ 
        CDMNet~\cite{duan2021coarse}  &FRCNN &ResNet50 &37,522 &16.8 &29.1 &18.5 &11.9 &29.0 &15.7  \\ 
        DREN~\cite{zhang2019fully}  &MaskRCNN+FPN &ResNet50 &-- &15.1 &-- &-- &-- &-- &-- \\
        DREN~\cite{zhang2019fully}  &MaskRCNN+FPN &ResNet101 &-- &17.1 &-- &-- &-- &-- &-- \\ 
        AMRNet~\cite{wei2020amrnet}  &FRCNN+FPN &ResNet50 &-- &18.2 &30.4 &19.8 &10.3 &31.3 &33.5 \\  

        GLSAN~\cite{deng2020global} &FRCNN+FPN &ResNet50 &-- &17.0 &28.1 &18.8 &-- &-- &-- \\
        GLSAN~\cite{deng2020global} &FRCNN+FPN &ResNet101 &-- &17.1 &28.3 &18.8 &-- &-- &-- \\ 
        CEASC~\cite{du2023ceasc} &GFL &-- &15,069 &17.1 &30.9 &17.8 &-- &-- &-- \\ 

        CenterNet~\cite{zhou2019objects} &-- &Hourglass104 &15,069 &13.2 &26.7 &11.8 &7.8 &26.6 &13.9  \\

        YOLC (Ours) &CenterNet &HRNet &30,138 &\textbf{19.3} &\textbf{30.9} &\textbf{20.1} &\textbf{10.9} &\textbf{32.2} &\textbf{35.5} \\ \bottomrule[1pt]
		\end{tabular}
	\end{center}
\label{tab:uavdt}
\end{table*}

\begin{table*}[!htbp]
  \caption{ The detection performance of each class on VisDrone~\cite{zhu2018visdrone} validation set. Ped. and Awn. are short for Pedestrian and Awning-tricycle. RS is short for random sampler.}
	\begin{center}
		\begin{tabular}{cc|cccccccccc}
        \midrule
        Method                         &Backbone &Ped. &Person &Bicycle &Car &Van &Truck &Tricycle &Awn. &Bus &Motor \\  \midrule 
                        \multicolumn{12}{c}{Comparison with base models}   \\  \midrule
        RetinaNet+RS~\cite{lin2017focal} &ResNet50 &13.0 &7.9 &1.4 &45.5 &19.9 &11.5 &6.3 &4.2 &17.8 &11.8\\
        FRCNN+RS~\cite{ren2016faster}  &ResNet50 &21.4 &15.6 &6.7 &51.7 &29.5 &19.0 &13.1 &7.7 &31.4 &20.7 \\
        FRCNN+RS~\cite{ren2016faster}  &ResNet101 &20.9 &14.8 &7.3 &51.0 &29.7 &19.5 &14.0 &8.8 &30.5 &21.2 \\
        FRCNN+RS~\cite{ren2016faster}  &ResNeXt101 &21.3 &15.5 &7.9 &53.0 &29.5 &20.5 &14.7 &8.9 &32.1 &21.6 \\
        CRCNN+RS~\cite{cai2018cascade} &ResNet50 &22.2 &14.8 &7.6 &54.6 &31.5 &21.6 &14.8 &8.6 &34.9 &21.4 \\ \midrule 

        RetinaNet+DSHNet~\cite{yu2021towards} &ResNet50 &14.1 &8.9 &1.3 &48.2 &24.8 &14.2 &8.8 &6.0 &21.6 &13.1 \\
        FRCNN+DSHNet~\cite{yu2021towards}  &ResNet50 &22.5 &16.5 &10.1 &52.8 &32.6 &22.1 &17.5 &8.8 &39.5 &23.7 \\
        FRCNN+DSHNet~\cite{yu2021towards}  &ResNet101 &21.7 &16.0 &10.1 &52.2 &31.6 &22.7 &17.1 &9.5 &38.6 &24.0 \\
        FRCNN+DSHNet~\cite{yu2021towards}  &ResNeXt101 &23.3 &16.7 &11.4 &53.7 &33.1 &23.8 &19.5 &11.1 &40.0 &25.5 \\
        CRCNN+DSHNet~\cite{yu2021towards}  &ResNet50 &23.2 &16.1 &11.2 &55.5 &33.5 &25.2 &19.1 &10.0 &43.0 &25.1 \\ \midrule \midrule
                      \multicolumn{12}{c}{Comparison with solutions to long-tail problems}  \\ \midrule
        FRCNN+RS+MMF~\cite{zhang2019dense_2} &ResNet50 &21.6 &15.3 &9.6 &51.5 &28.5 &20.4 &15.9 &7.5 &33.7 &21.6 \\
        FRCNN+SimCal~\cite{wang2020devil}  &ResNet50 &18.7 &13.8 &5.7 &51.0 &28.4 &16.4 &13.6 &5.9 &27.0 &19.4 \\
        FRCNN+RS+BGS~\cite{li2020overcoming} &ResNet50 &21.8 &16.0 &8.1 &51.8 &31.1 &19.8 &15.0 &8.4 &36.1 &21.5 \\ \midrule
        FRCNN+DSHNet~\cite{yu2021towards}  &ResNet50 &22.5 &16.5 &10.1 &52.8 &32.6 &22.1 &17.5 &8.8 &39.5 &23.7 \\   \midrule \midrule
                      \multicolumn{11}{c}{Based on the SOTA cropping method}   \\ \midrule
        DMNet (FRCNN+RS)~\cite{li2020density}  &ResNet50 &28.1 &19.7 &13.3 &57.3 &36.1 &24.8 &20.1 &12.0 &42.9 &26.4 \\  \midrule
        DMNet~\cite{li2020density}cropping+DSHNet~\cite{yu2021towards}  &ResNet50 &28.5 &20.4 &15.9 &56.8 &37.9 &30.1 &22.6 &14.0 &47.1 &29.2 \\ \midrule 

        CenterNet~\cite{zhou2019objects} (Baseline) &Hourglass104 &28.7 &18.5 &12.7 &57.9 &37.6 &29.6 &17.4 &11.4 &43.2 &20.9 \\
        YOLC (k=0) &HRNet &37.4 &24.3 &21.3 &64.3 &43.8 &34.0 &26.5 &17.9 &53.2 &33.6 \\
        YOLC (k=1) &HRNet &39.2 &26.3 &22.9 &\textbf{65.5} &\textbf{45.2} &36.0 &\textbf{28.9} &\textbf{19.7} &\textbf{58.9} &\textbf{35.6} \\
        YOLC (k=2) &HRNet &39.4 &\textbf{26.5} &23.5 &65.4 &\textbf{45.2} &\textbf{36.5} &28.6 &\textbf{19.7} &58.7 &35.5 \\
        YOLC (k=3) &HRNet &\textbf{39.5} &\textbf{26.5} &\textbf{23.6} &65.3 &\textbf{45.2} &\textbf{36.5} &28.6 &\textbf{19.7} &58.8 &35.5  \\ \hline

		\end{tabular}
	\end{center}
\label{tab:single}
\end{table*}

The performance evaluation on the UAVDT dataset, as shown in Table~\ref{tab:uavdt}, reveals a similar conclusion to the VisDrone~\cite{zhu2018visdrone} dataset. It demonstrates that general object detectors are unable to achieve satisfactory detection results, whereas our proposed YOLC outperforms state-of-the-art models and achieves the highest performance with an AP of 19.3. Notably, YOLC consistently improves the accuracy of small, medium, and large objects, which validates the effectiveness of our dedicated detection framework.

\begin{figure*}[!htbp]
	\centering
	\includegraphics[width=1.0 \linewidth]{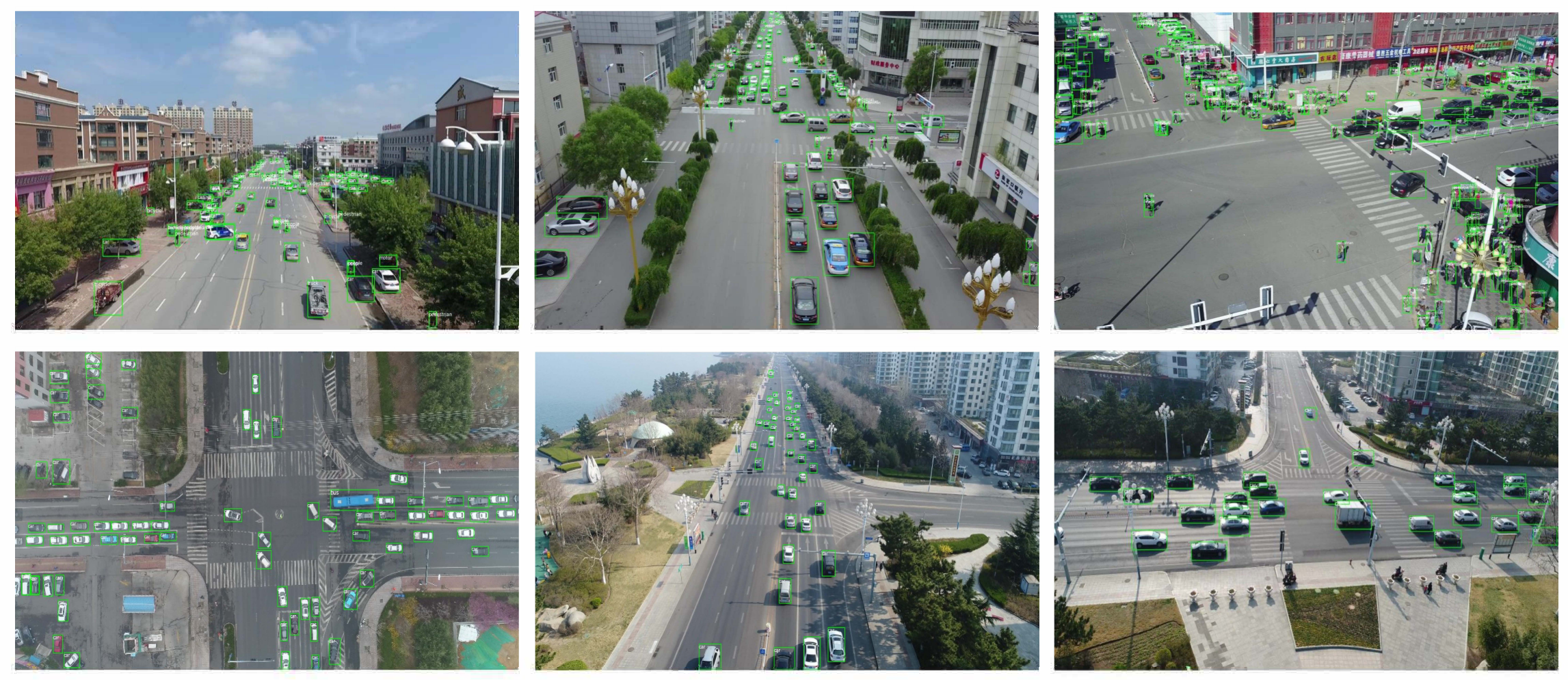}
	\caption{Visualization of our YOLC detection results on VisDrone~\cite{zhu2018visdrone} (first row) and UAVDT~\cite{du2018unmanned} (second row).}
	\label{fig:detection}
\end{figure*}

\begin{figure*}[!htbp]
	\centering
	\includegraphics[width=1.0 \linewidth]{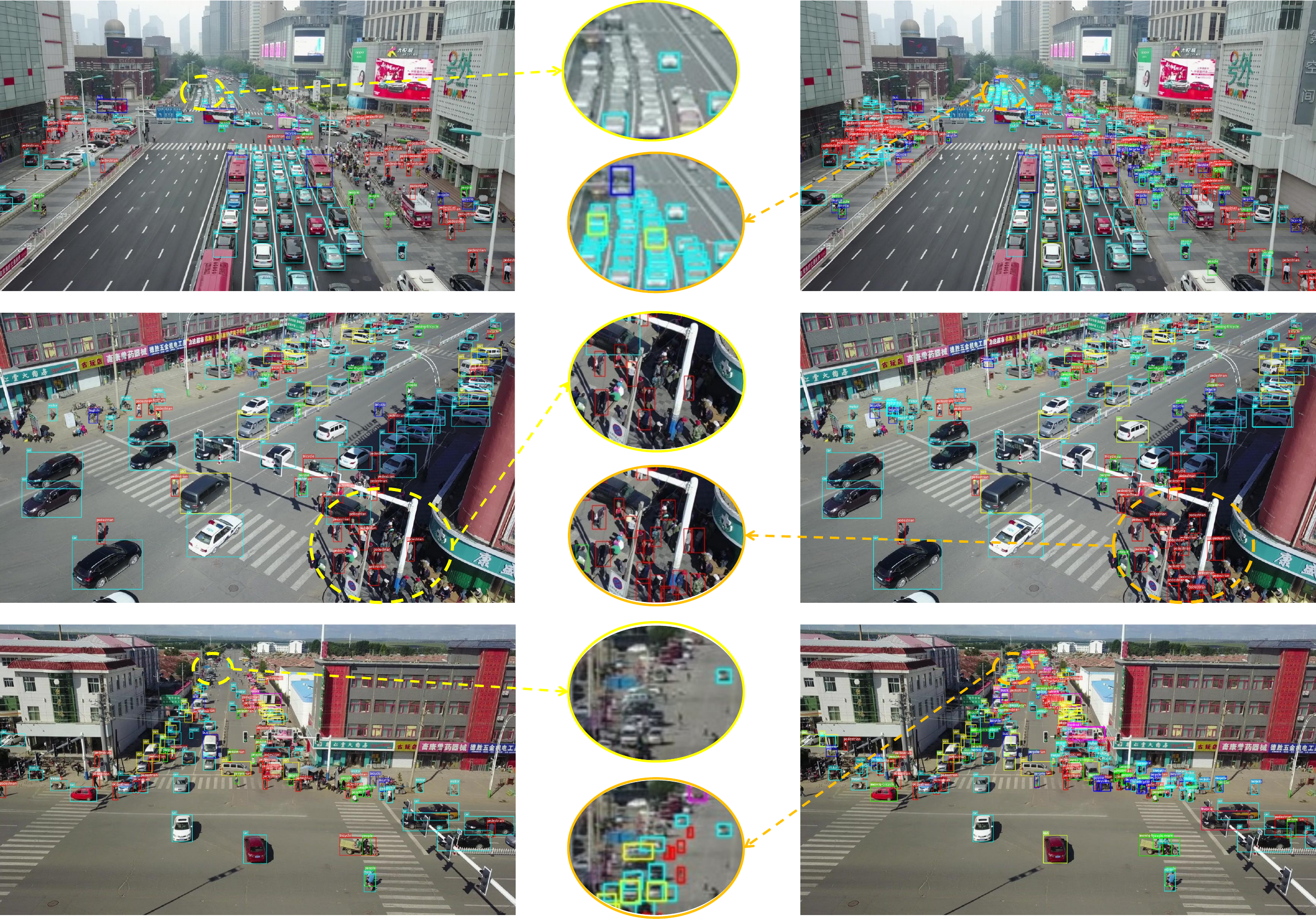}
	\caption{Visualization of our YOLC (right) and CenterNet (left) detection results on VisDrone.}
	\label{fig:comp}
\end{figure*}

We further analyze the class-wise APs on VisDrone~\cite{zhu2018visdrone} in Table~\ref{tab:single}. Our proposed YOLC consistently outperforms the base models, including RetinaNet~\cite{lin2017focal}, Faster R-CNN~\cite{ren2016faster}, and Cascade R-CNN~\cite{cai2018cascade}, by a large margin, particularly for small objects like pedestrians and persons. However, we observe that some categories, such as car and bus, achieve better detection performance compared to others, such as bicycle and awning-tricycle, which can be attributed to severe class imbalance in the dataset. Several approaches, including re-sampling~\cite{buda2018systematic,byrd2019effect}, re-weighting~\cite{cui2019class}, re-balancing~\cite{yang2020rethinking,zhou2020bbn}, and two-phase training paradigms~\cite{wang2020devil,yu2021towards}, have been proposed to address this issue. Despite not focusing on the long-tail problem, our method still achieves the highest performance across all categories compared to several models designed for long-tail problems, as shown in Table~\ref{tab:single}.

Moreover, we visualize the detection results of both datasets in Fig.~\ref{fig:detection}. From these visualizations, we can observe that our proposed method achieves promising detection accuracy. In particular, our method can detect many small objects that are easily missed by general detectors, as shown in the images in the second column. Additionally, our method can obtain accurate detection results for non-uniformly distributed objects, as shown in the third image in the first row. Furthermore, we visualize the results of YOLC compared with CenterNet in Fig.~\ref{fig:comp}. YOLC can detect objects more accurately in dense regions than CenterNet, especially small objects.

\subsection{Ablation Study}
To demonstrate the effectiveness of the components in our model, we conduct ablation experiments on VisDrone~\cite{zhu2018visdrone}. In our framework, LSM serves as the critical component, exhibiting the most significant im-provement in overall performance.

\noindent $\bullet$ \textbf{Different backbones.}
To enhance the detection of small objects in aerial images, we have replaced the Hourglass backbone network in CenterNet~\cite{zhou2019objects} with a high-resolution network, HRNet~\cite{sun2019high}. Table~\ref{tab:visdrone} shows that this simple modification leads to a 1.3\% improvement in performance, demonstrating the effectiveness of using HRNet to extract more robust features, particularly for small objects. Therefore, in our subsequent experiments, we use HRNet as our backbone network.

\noindent $\bullet$ \textbf{Effect of LSM.} 
We propose an LSM module to address the non-uniform distribution problem. As reported in Table~\ref{tab:visdrone}, we observe that the performance improves by 2.2\% when the number of cropped clustered regions is set to $k=1$. It is noteworthy that the performance gain for small objects is even more significant, with an improvement of 3.3\%. However, as $k$ increases, the performance improvement saturates. With $k=2$ and $k=3$, we observe a further improvement of 0.3\% and 0.1\%, respectively, on $AP_{small}$. Therefore, to fully leverage the benefits of the LSM module, we set $k=3$ in our subsequent experiments.

\begin{figure*}[!htbp]
	\centering
	\includegraphics[width=1.0 \linewidth]{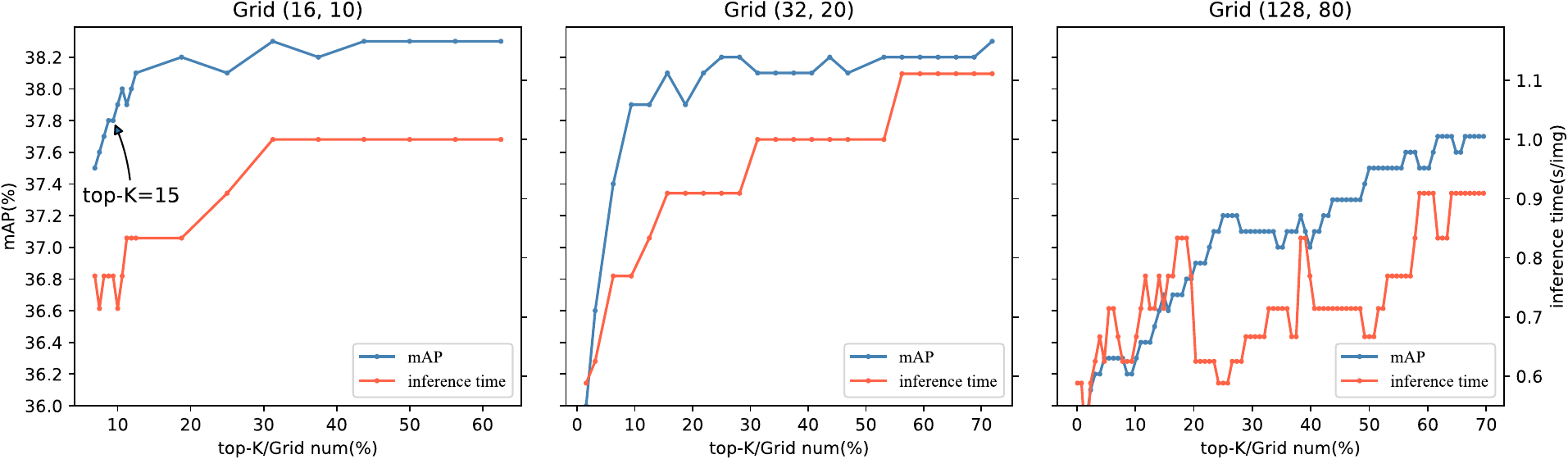}
	\caption{The performance and efficiency with different Grid size and top-K. Grid size is set to (16, 10), (32, 20), and (128, 80) respectively. X-axis represents top-K divided by Grid number as a percentage. Note that we show the original hyperparameters setup (Grid= (16, 10), top-K=15).}
	\label{fig:hyper}
\end{figure*}

\noindent $\bullet$ \textbf{Hyperparameters in LSM.} 
We present the ablation of two hyperparameters Grid and top-K in LSM (see Fig.~\ref{fig:hyper}). Grid denotes the grids' size and top-K denotes the number of candidate regions. Specially, we observed that the initial hyperparameter configuration is not optimal. When Grid is (16, 10) and top-K is above 30\% of the total number of Grid, optimal results are achieved, reaching 38.3\% mAP.

\noindent $\bullet$ \textbf{Different loss functions.} 
To optimize the proposed YOLC network, we made several changes to the regression loss. Specifically, we replaced the original $L_{1}$ loss with the GWD loss~\cite{yang2021rethinking} and experimented with IoU-based losses such as GIoU~\cite{rezatofighi2019generalized}, DIoU~\cite{zheng2020distance}, and CIoU~\cite{zheng2020distance}. We also used the modified GIoU loss to adapt our anchor-free detector~\cite{gong2020towards}. These losses were chosen to address the sensitivity to scale variation and the inconsistency between metric and loss. The results of these experiments are shown in Table~\ref{tab:loss}. We observed further improvements in performance by 0.4\% and 0.8\% AP when using the alternative regression losses, with $AP_{small}$ showing improvements of 0.5\% and 0.9\%. However, $AP_{medium}$ and $AP_{large}$ decreased. Based on the analysis in Section~\ref{subsec:loss}, we found that the GWD+$L_1$ loss alleviated this phenomenon, with $AP_{small}$ showing further improvements by 1.3\%.

\noindent $\bullet$ \textbf{Effect of refinement module.} 
To further enhance the performance of the YOLC network on small objects, we incorporate a deformable convolution module to refine the regression head. This module can dynamically extract point features to generate a more precise representation for each object. Table~\ref{tab:refine} demonstrates that the proposed deformable convolution improves $AP_{75}$ by 1.4\%, indicating that bounding box regression achieves higher localization accuracy. Furthermore, all evaluation metrics show consistent improvements, with larger gains observed for larger objects. This improvement can be attributed to the ability of deformable convolution to better capture the entire object, especially for large objects where traditional convolution may struggle to cover the object's entirety.

\noindent $\bullet$ \textbf{Effect of decoupled heatmap branch.} 
To improve object localization accuracy, we disentangle the heatmap branch into parallel branches, where each branch independently generates a heatmap for the corresponding object category. The results in Table~\ref{tab:decoupled} demonstrate that this approach yields consistent improvement across all evaluation metrics, validating the effectiveness of the decoupled heatmap. Notably, $AP_{small}$ is further improved by 0.4\%, indicating that this technique is particularly effective for detecting small objects.

\setlength{\tabcolsep}{3pt}

\begin{table}[!htbp]
        \vspace{-0.3cm}
        \caption{Performance comparison on different losses.}
              \begin{center}
                      \begin{tabular*}{\linewidth}{c|ccc|ccc}
              \midrule
              Method  &$AP$ &$AP_{50}$ &$AP_{75}$ &$AP_{\rm small}$ &$AP_{\rm medium}$ &$AP_{\rm large}$ \\  \midrule
      
              LSM+baseline ($L_1$) &33.0 &58.2 &33.1 &26.8 &43.2 &50.1  \\
      
              LSM+GIoU &33.4 &58.0 &33.8 &27.3 &43.3 &46.2  \\
              LSM+GWD &33.8 &58.3 &34.3 &27.7 &44.1 &47.3  \\ 
              LSM+GWD+$L_1$ &34.2 &58.5 &34.6 &28.1 &44.2 &48.2  \\  \midrule
                      \end{tabular*}
              \end{center}
\label{tab:loss}
\end{table}

\setlength{\tabcolsep}{5pt}

\begin{table}[!htbp]
        \vspace{-0.3cm}
        \caption{Performance comparison on the refine module.}
              \begin{center}
                      \begin{tabular*}{\linewidth}{c|ccc|ccc}
              \midrule
              Method  &$AP$ &$AP_{50}$ &$AP_{75}$ &$AP_{\rm small}$ &$AP_{\rm medium}$ &$AP_{\rm large}$ \\  \midrule
      
              w/o refine      &33.8 &58.3 &34.3 &27.7 &44.1 &47.3  \\
              w/ refine &34.6 &58.6 &35.7 &28.0 &45.3 &51.1  \\  \midrule
                      \end{tabular*}
              \end{center}
\label{tab:refine}
\end{table}

\begin{table}[!htbp]
        \vspace{-0.3cm}
        \caption{Performance comparison on the decoupled heatmap.}
              \begin{center}
                      \begin{tabular*}{\linewidth}{c|ccc|ccc}
              \midrule
              Method  &$AP$ &$AP_{50}$ &$AP_{75}$ &$AP_{\rm small}$ &$AP_{\rm medium}$ &$AP_{\rm large}$ \\  \midrule
      
              coupled      &34.6 &58.6 &35.7 &28.0 &45.3 &51.1  \\
              decoupled    &35.0 &59.3 &36.2 &28.5 &45.6 &50.9  \\  \midrule
                      \end{tabular*}
              \end{center}
\label{tab:decoupled}
\end{table}

\section{Conclusions}
This paper introduces the YOLC framework, an anchor-free network designed to tackle two challenging problems in aerial object detection: detecting small objects and dealing with severely non-uniform data distribution. To address the former issue, we present an improved model based on CenterNet that uses transpose convolution to output high-resolution feature maps and decouples heatmaps to learn dedicated representations for different object categories. This approach leads to significant improvements in detection performance, particularly for small objects. For the latter problem, we introduce a local scale module that adaptively searches for clustered object regions and rescales them to better suit the object detectors. To further improve the performance of the network, we adopt GWD-based loss functions to replace the size regression loss in the original CenterNet, which focuses the model on small objects. To compensate for the drop in performance of large objects caused by the above loss, we propose using $L_1$ loss to assist GWD loss. Additionally, we utilize a deformable module to refine bounding box regression. Our framework balances accuracy and speed through the LSM. We demonstrate the effectiveness and superiority of our approach through extensive experiments conducted on two aerial image datasets compared to state-of-the-art methods.

In future work, we will endeavor to expand our method to feature level for tiny object detection. For instance, the model can directly learn the features of upscaled images, as opposed to inputting upscaled images into the backbone network to obtain features.

\bibliographystyle{IEEEtran}
\bibliography{references}

\begin{IEEEbiography}
[{\includegraphics[width=1in,height=1.25in,clip,keepaspectratio]{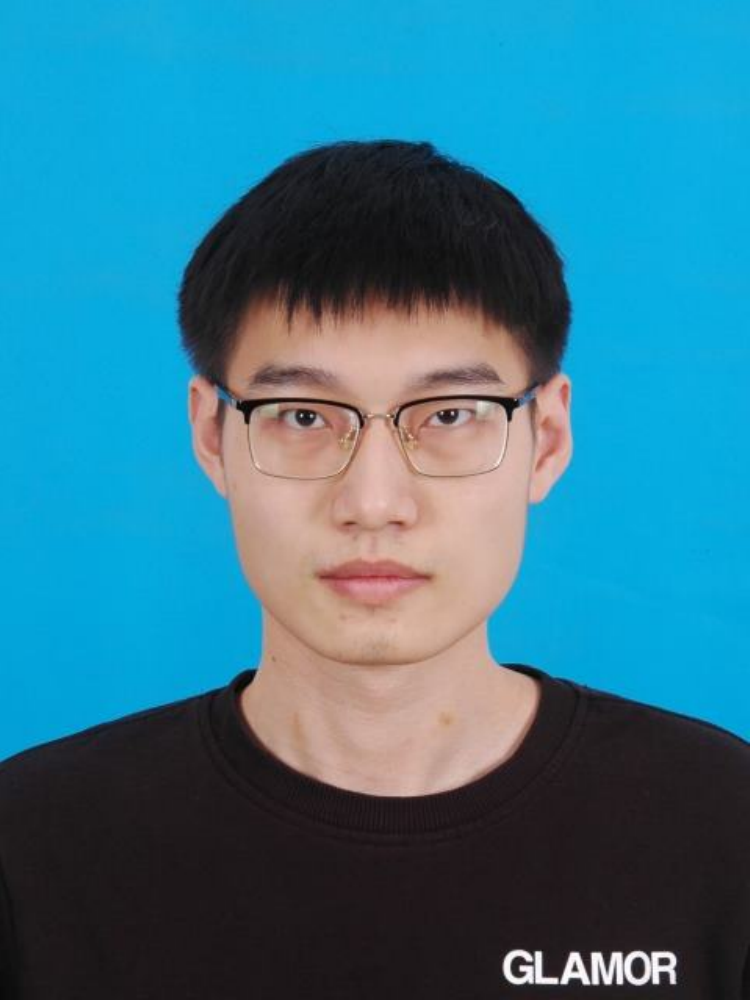}}]{Chenguang Liu}
received the B.S. degree in computer science from the School of Computer Science and Engineering, Beihang University, Beijing, China, in 2022. He is currently pursuing the M.S. degree in software engineering with the Laboratory of Intelligent Recognition and Image Processing, School of Computer Science and Engineering, Beihang University. His research interests include computer vision and pattern recognition.
\end{IEEEbiography}

\begin{IEEEbiography}
[{\includegraphics[width=1in,height=1.25in,clip,keepaspectratio]{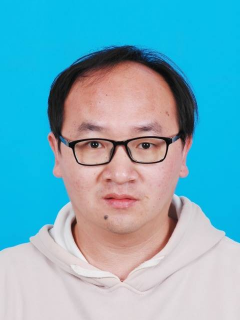}}]{Guangshuai Gao}
received the B.S. degree in applied physics from college of science and the M.S. degree in signal and information processing from the School of Electronic and Information, from the Zhongyuan University of Technology, Zhengzhou, China, in 2014 and 2017, respectively. He received the Ph.D degree with the Laboratory of Intelligent Recognition and Image Processing, Beijing Key Laboratory of Digital Media, School of Computer Science and Engineering, Beihang University, Beijing, China, in 2022.

He is currently a lecturer with the School of Electronic and Information Engineering, Zhongyuan University of Technology. His research interests include image processing, pattern recognition, remote sensing image analysis and digital machine learning.
\end{IEEEbiography}

\begin{IEEEbiography}
[{\includegraphics[width=1in,height=1.25in,clip,keepaspectratio]{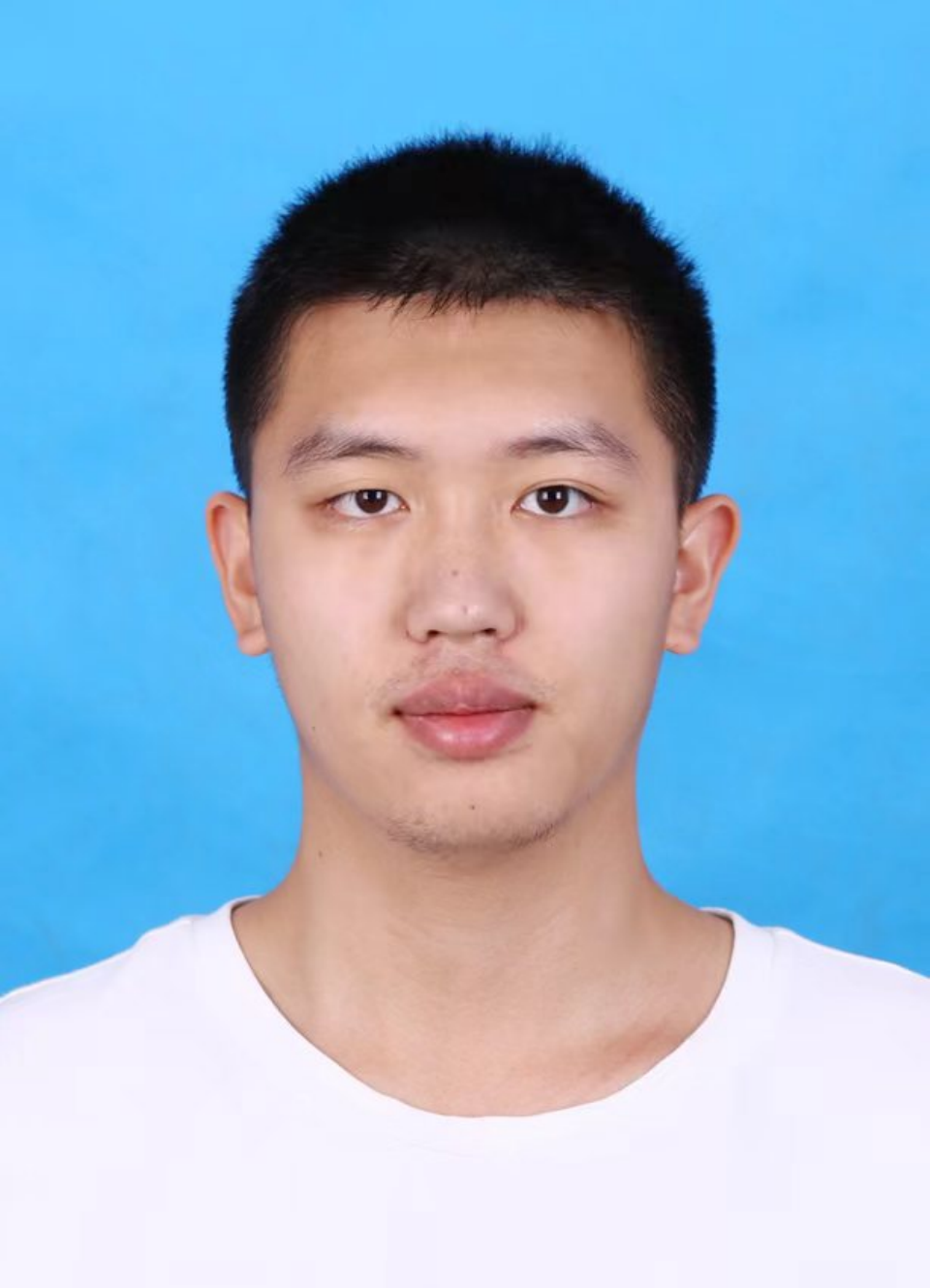}}]{Ziyue Huang}
received the BS degree in information and computing science from the School of Mathematics and Statistics, Beijing Institute of Technology, Beijing, China, in 2018. He is currently pursuing the Ph.D. degree in computer science with the Laboratory of Intelligent Recognition and Image Processing, School of Computer Science and Engineering, Beihang University.

His research interests include computer vision and object detection.
\end{IEEEbiography}

\begin{IEEEbiography}
[{\includegraphics[width=1in,height=1.25in,clip,keepaspectratio]{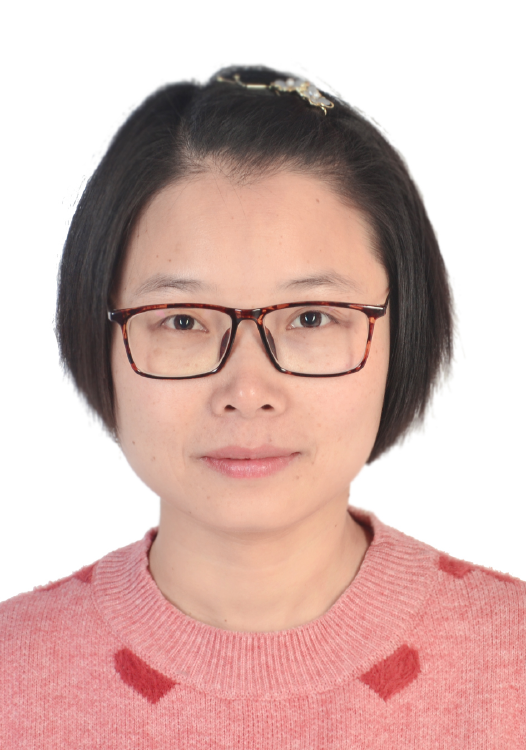}}]{Zhenghui Hu}
received the B.S. degree in computer science from the Zhejiang University of Technology, Hangzhou, China, in 2011, and the Ph.D. degree in computer science from Beihang University, Beijing, China, in 2020.

She is currently a Senior Research Associate with Hangzhou Innovation Institute, Beihang University. Her research interests include computer vision, hybrid intelligence, and crowdsourcing-based software engineering.
\end{IEEEbiography}

\newpage

\begin{IEEEbiography}
[{\includegraphics[width=1in,height=1.25in,clip,keepaspectratio]{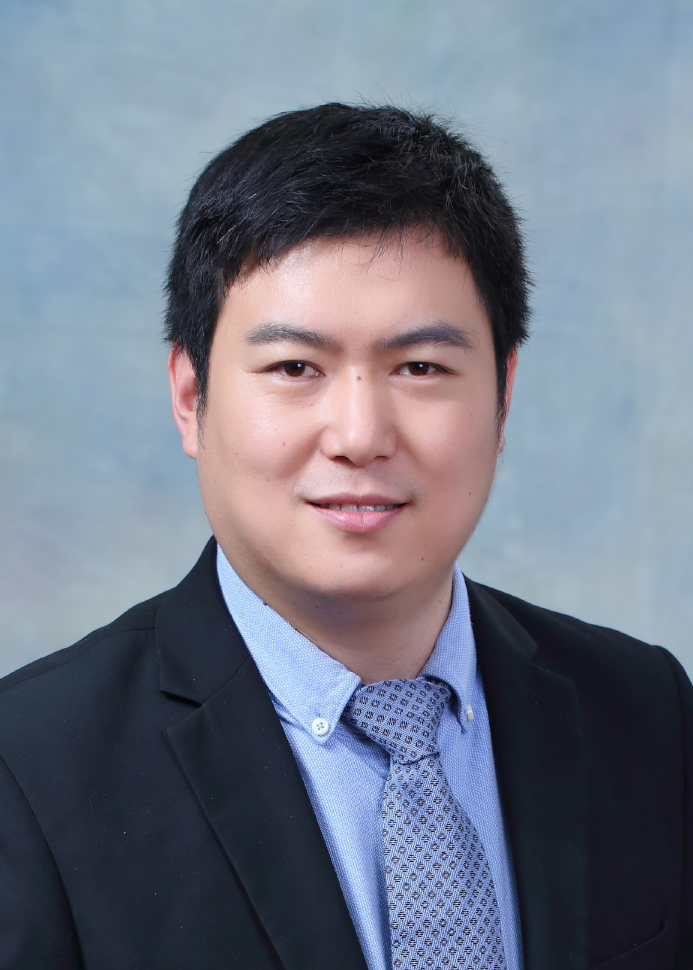}}]{Qingjie Liu}
received the BS degree in computer science from Hunan University, Changsha, China, in 2007, and the Ph.D. degree in computer science from Beihang University, Beijing, China, in 2014. 

He is currently an Associate Professor with the School of Computer Science and Engineering, Beihang University. He is also a Distinguished Research Fellow with the Hangzhou Institute of Innovation, Beihang University, Hangzhou. His current research interests include image fusion, object detection, image segmentation, and change detection. He is a member of the IEEE.
\end{IEEEbiography}

\begin{IEEEbiography}
[{\includegraphics[width=1in,height=1.25in,clip,keepaspectratio]{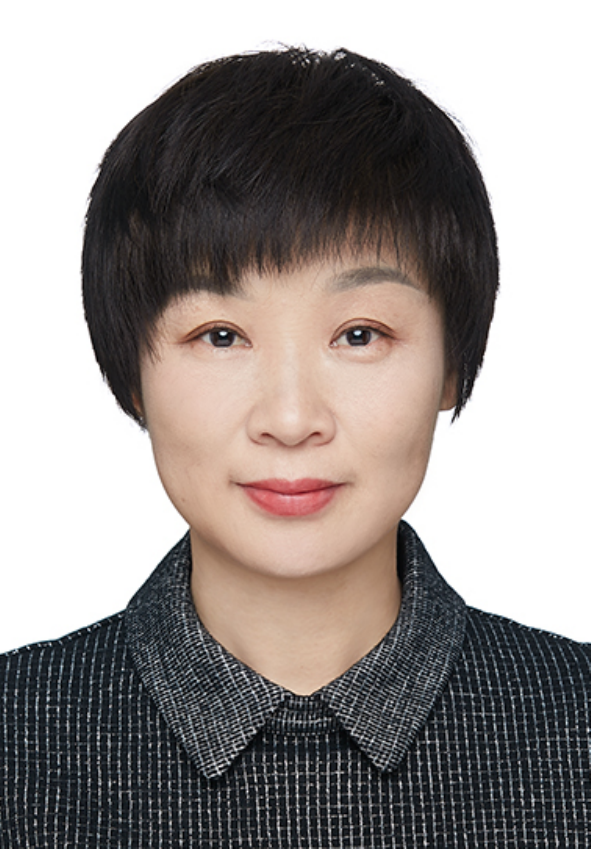}}]{Yunhong Wang}
received the BS degree in electronic engineering from Northwestern Polytechnical University, Xi'an, China, in 1989, and the MS and Ph.D. degrees in electronic engineering from the Nanjing University of Science and Technology, Nanjing, China, in 1995 and 1998, respectively.

She was with the National Laboratory of Pattern Recognition, Institute of Automation, Chinese Academy of Sciences, Beijing, China, from 1998 to 2004. Since 2004, she has been a Professor with the School of Computer Science and Engineering, Beihang University, Beijing, where she is also the Director of the Laboratory of Intelligent Recognition and Image Processing. Her research interests include biometrics, pattern recognition, computer vision, data fusion, and image processing.

Dr. Wang is a Fellow of IEEE, IAPR, and CCF.
\end{IEEEbiography}

\enlargethispage{-6.5cm}

\newpage
\vfill
\end{document}